\def\year{2022}\relax
\documentclass[letterpaper]{article} 
\usepackage{aaai22}  
\usepackage{times}  
\usepackage{helvet}  
\usepackage{courier}  
\usepackage[hyphens]{url}  
\usepackage{graphics} 
\usepackage{graphicx}\graphicspath{ {./figures/} }
\usepackage{amsmath} 
\usepackage{amssymb}  
\usepackage{bm}
\usepackage{multicol}
\usepackage{booktabs}
\usepackage{xcolor}
\usepackage{subcaption}
\usepackage{mathtools}
\usepackage{amsthm}
\usepackage{array}
\newcolumntype{L}{>{\raggedright\arraybackslash}p{2cm}}
\newtheorem{defi}{Definition}
\newtheorem{theo}{Theorem}

\DeclareMathOperator*{\argmax}{arg\,max}

\DeclarePairedDelimiter\floor{\lfloor}{\rfloor}
\DeclareMathSizes{6.5}{6.5}{6.5}{6.5}

\usepackage[switch]{lineno}  %
\usepackage[para]{footmisc}
\usepackage{cite}
\usepackage{istgame}
\usepackage{subcaption}
\usepackage{amsmath}
\usepackage{ upgreek }
\usepackage{multirow}
\usepackage{textcomp}
\usepackage{siunitx}
\usepackage{adjustbox}
\usepackage{bm}
\usepackage{tikz}
\usetikzlibrary{automata,positioning}
\usetikzlibrary{decorations.pathmorphing}
\newcommand\addvmargin[1]{
}
\usepackage{accents}

\newcolumntype{P}[1]{>{\centering\arraybackslash}p{#1}}

\usepackage{natbib}  
\usepackage{caption}[font=small,skip=2pt] 
\DeclareCaptionStyle{ruled}{labelfont=normalfont,labelsep=colon,strut=off} 
\title{Generalized dynamic cognitive hierarchy models for strategic driving behavior}

\author{
    Atrisha Sarkar \textsuperscript{\rm 1},
    Kate Larson \textsuperscript{\rm 1},
    Krzysztof Czarnecki \textsuperscript{\rm 2}
    \\
}
\affiliations{
\textsuperscript{\rm 1}David R Cheriton School of Computer Science
\textsuperscript{\rm 2}Department of Electrical and Computer Engineering\\
University of Waterloo, Ontario, Canada\\

    atrisha.sarkar, kate.larson, k2czarne@uwaterloo.ca

}
\begin{document}
\maketitle
\begin{abstract}
While there has been an increasing focus on the use of game theoretic models for autonomous driving, empirical evidence shows that there are still open questions around dealing with the challenges of common knowledge assumptions as well as modeling bounded rationality. To address some of these practical challenges, we develop a framework of generalized dynamic cognitive hierarchy for both modelling naturalistic human driving behavior as well as behavior planning for autonomous vehicles (AV). This framework is built upon a rich model of level-0 behavior through the use of automata strategies, an interpretable notion of bounded rationality through safety and maneuver satisficing, and a robust response for planning. Based on evaluation on two large naturalistic datasets as well as simulation of critical traffic scenarios, we show that i) automata strategies are well suited for level-0 behavior in a dynamic level-k framework, and ii) the proposed robust response to a heterogeneous population of strategic and non-strategic reasoners can be an effective approach for game theoretic planning in AV.
\end{abstract}

\section{Introduction}
One of many challenges of autonomous vehicles (AV) in an urban setting is the safe handling of other human road users who show complex and varied driving behaviors. As AVs integrate into human traffic, there has been a move from a \emph{predict-and-plan} approach of behavior planning to a more strategic approach where the problem of behavior planning of an AV is set up as a non-zero sum game between road users and the AV; and such models have shown efficacy in simulation of different traffic scenarios \cite{fisac2019hierarchical, tian2018adaptive, li2019decision, sadigh2016planning}. However, when game theoretic models are evaluated on naturalistic human driving data, studies have shown that human driving behavior is diverse, and therefore developing a unifying framework that can both model the diversity of human driving behavior as well as plan a response from the perspective of an AV is challenging \cite{sun2020game, sarkar2021solution}. \par
The first challenge is dealing with \emph{common knowledge} \cite{geanakoplos1992common} assumptions. Whether in the case of Nash equilibiria based models (rational agents assuming everyone else is a rational agent) \cite{schwarting2019social, Geiger_Straehle_2021}, in Stackelberg equilibrium based models (common understanding of the leader-follower relationship) \cite{fisac2019hierarchical}, or level-k model (the level of reasoning of AVs and humans), there has to be a consensus between an AV planner and other road users on the type of reasoning everyone is engaging in. How to reconcile this assumption with the evidence that human drivers engage in different types of reasoning processes is a key question. \par
Whereas the challenge of common knowledge deals with questions around how agents model other agents, the second challenge is dealing with bounded rational agents, i.e. modelling sub-optimal behavior in their own response. With the general understanding that human drivers are bounded rational agents, the Quantal level-k model (QLk) has been widely proposed as a model of strategic planning for AV \cite{tian2018adaptive, li2018game, li2019decision}. One way the QLk model deals with bounded rationality is through suboptimal reasoning, i.e., by mapping each agent into a hierarchy ($k$) of bounded cognitive reasoning. In this model, the choice of level-0 model becomes critical, since the behavior of every agent in the hierarchy depends on the assumption about the behavior of level-0 agents. The main models proposed for level-0 behavior include simple obstacle avoidance \cite{tian2021anytime}, maxmax, and maxmin models \cite{wright2020formal, sarkar2021solution}. Although such elementary models may be acceptable in other domains of application, it is not clear why human drivers, especially in a dynamic game setting, would cognitively bound themselves to such elementary models. Another way QLk model deals with bounded rationality is through suboptimal response, i.e., by the use of a precision parameter, where agents instead of selecting the utility maximizing response, make cost proportional errors modelled by the precision parameter \cite{wright2010beyond}. While the precision parameter provides a convenient way to fit the model to empirical data, it also runs the risk of being a catch-all for different etiologies of bounded rationality, including, people being indifferent to choices, uncertainty around multiple objectives, as well as the model not being a true reflection of the decision process. This impairs the explainability of the model since it is hard to encode and communicate such disparate origins of bounded rationality through a single parameter.\par
The primary contribution of our work is a framework that addresses the aforementioned challenges by unifying modeling of heterogeneous human driving behavior with strategic planning for AV. In this framework, behavior models are mapped into three layers of increasing capacity to reason about other agents' behavior -- \emph{non-strategic}, \emph{strategic}, and \emph{robust}. Within each layer, the possibility of different types of behavior models lends support for a population of heterogeneous behavior, with a robust layer on top addressing the problem of behavior planning with a relaxed common knowledge assumptions. Standard level-k type and equilibrium models are nested within this framework, and in the context of those models, secondary contributions of our work are a) the use of automata strategies as a model of level-0 behavior in dynamic games, resulting in behavior that is rich enough to capture naturalistic human driving (dLk($\mathcal{A}$) model), and b) an interpretable support for bounded rationality based on different modalities of satisficing --- \emph{safety} and \emph{maneuver}. Finally, the efficacy of the approach is demonstrated with evaluation on two large naturalistic driving datasets as well as simulation of critical traffic scenarios.\par

\section{Game tree, utilities, and agent types}
\label{sec:game_tree}
\begin{figure}[t]
  \centering
    \includegraphics[width=.25\linewidth]{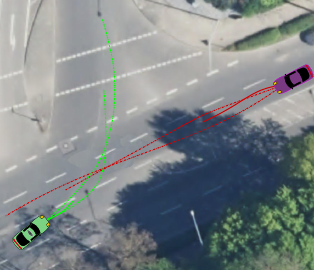}
    \includegraphics[width=.5\linewidth]{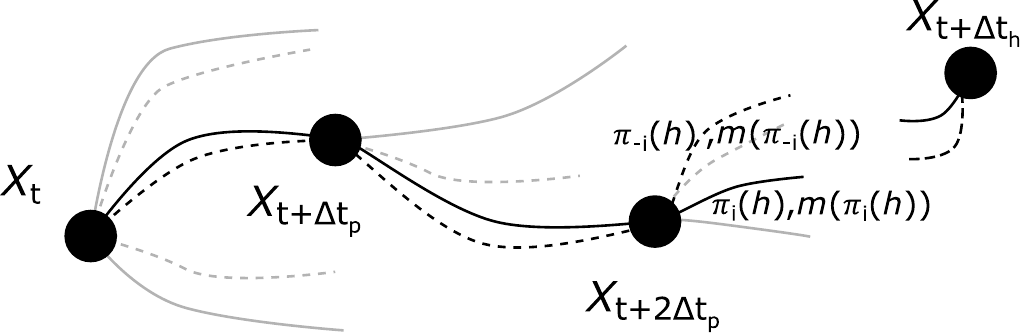}
    \caption{Schematic representation of the dynamic game. Each node is embedded in a spatio-temporal lattice and nodes are connected with a cubic spline trajectory.}
  \label{fig:game_tree}
\end{figure}
The dynamic game is constructed as a sequence of simultaneous move games played starting at time $t=0$ at a period of $\Delta t_{p}$ secs. over a horizon of $\Delta t_{h}$ secs. Each vehicle $i \in \{1,2,..,N\}$'s state at time $t$ is a vector $X_{i,t} = [x,y,v_{x},v_{y},\dot{v_x},\dot{v_y},\theta]$ representing positional co-ordinates ($x,y$) on $R^{2}$, lateral and longitudinal velocity ($v_{x},v_{y}$) in the body frame, acceleration ($\dot{v_x},\dot{v_y}$), and yaw ($\theta$). The nodes of the game tree $X_{t} = X_{i,t}^N$ are the joint system states embedded in a spatio-temporal lattice \cite{ziegler2009spatiotemporal}, and the actions are cubic spline trajectories \cite{kelly2003reactive, al2018spline} generated based on kinematic limits of vehicles with respect to bounds on lateral and longitudinal velocity, acceleration, and jerk \cite{bae2019toward}. A history $h_t$ of the game consists of sequence of nodes $X_{0}..X_{t}$ traversed by both agents along the game tree until time $t$. We also use a hierarchical approach in the trajectory generation process \cite{michon1985critical, fisac2019hierarchical, sarkar2021solution}, where at each node, the trajectories are generated with respect to high-level maneuvers, namely, \emph{wait} and \emph{proceed} maneuvers. For wait maneuver trajectories, a moving vehicle decelerates (or remain stopped if it is already stopped), and for proceed maneuvers, a vehicle maintains its moving velocity or accelerates to a range of target speeds. Strategies are presented in the behavior strategy form, where $\pi_{i}(h_{t}) \in \mathrm{T_{i}(X_{t})}$ is a pure strategy response (a trajectory) of an agent $i$ that maps a history $h_{t}$ to a trajectory in the set $\mathrm{T_{i}(X_{t})}$, which is the set of valid trajectories that can be generated at the node $X_{t}$ corresponding to both maneuvers. The associated maneuver for a trajectory is represented as $m(\pi_{i}(h)) \in \{\mathit{wait,proceed}\}$. Depending on the context where the response depends on only the current node instead of the entire history, we use the notation $\pi_{i}(X_{t})$; we also drop the time subscript $t$ on the history when a formulation holds true for all $t$. The overall strategy of the dynamic game is the cross product of behavior strategies along all possible histories of the game $\sigma : \pi_{1}(h) \times \pi_{2}(h) \times .. \pi_{N}(h); \forall h$. We use the standard game-theoretic notation of $i$ and $-i$ to refer to an agent and other agents respectively in a game. \par
The utilities in the game are  formulated as multi-objective utilities consisting of two components --- \textbf{safety} $u_{s,i}(\pi_{i}(h),\pi_{-i}(h)) \in [-1,1]$ (modelled as a sigmoidal function that maps the minimum distance gap between trajectories to a utility interval [-1,1]) and \textbf{progress} $u_{p,i}(\pi_{i}(h),\pi_{-i}(h)) \in [0,1]; \forall i, -i$ (a function that maps the trajectory length in meters to a utility interval [0,1]). In general, these two are the main utilities (often referred to as inhibitory and excitatory utilities, respectively) upon which different driving styles are built \cite{sagberg2015review}. Agent types $\gamma_{i} \in \Gamma$ are numeric in the range [-1,1] representing each agent's \emph{safety aspiration level} --- a level of safety that an agent is comfortable operating. This construction is motivated by traffic behavior models such as Risk Monitoring Model \cite{vaa2011drivers}, Task Difficulty Homeostasis theory \cite{fuller2008recent}, where based on traffic situations, drivers continually compare each possible action to their own risk tolerance threshold and make decisions accordingly. To avoid dealing with complexities that arise out of agent types being continuous, for the scope of this paper, we discretize the types by increments of 0.5 for our experiments. Based on their type (i.e. their safety aspiration level) how each agent selects specific actions at each node depends on the particular behavior model, and is elaborated in Sec. \ref{sec:main} when we discuss the specifics of each behavior model. \par
Unless mentioned otherwise, the utilities (both safety and progress) at a node with associated history $h_{t}$ are calculated as discounted sum of utilities over the horizon of the game conditioned on the strategy $\sigma$, type $\gamma_{i}$ and discount factor $\delta$ as follows
\begin{linenomath*}
\begin{align*}
    \sum\limits_{k=1}^{\floor*{\Delta t_{h}-t \backslash \Delta t_{p} }} \delta^{k}u_{i}(\pi_{i}(h_{t+k-1}),\pi_{-i}(h_{t+k-1});\sigma,\gamma_{i}) + \mathcal{N}u_{i,C}
\end{align*}
\end{linenomath*}
where $u_{i,C}$ are the continuation utilities beyond the horizon of the game and are estimated based on agents continuing on with the same chosen trajectory as undertaken in the last decision node of the game tree for another $\Delta t_{h}$ seconds, and $\mathcal{N}$ is a normalization constant to keep the sum of utilities in the same range as the step utilites.
\section{Generalized dynamic cognitive hierarchy model}
\label{sec:main}
\begin{figure}[h]
  \centering
    \includegraphics[width=.45\linewidth]{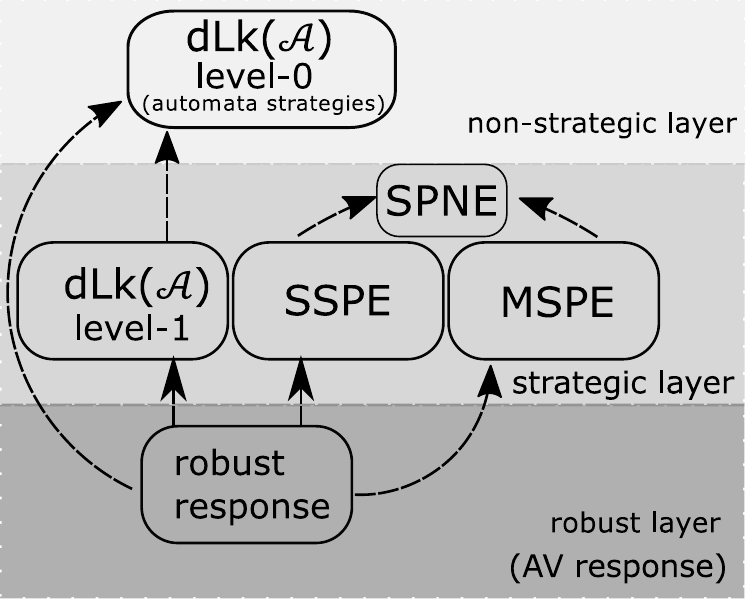}
    \includegraphics[width=.45\linewidth]{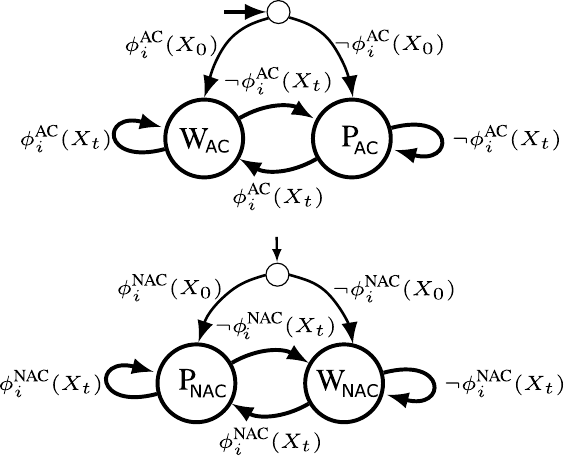}
    \caption{(a) Organization of models in the generalized dynamic cognitive hierarchy framework. Dashed arrows indicate agents' belief about the population. (b) Automata $\mathcal{A}^{\text{AC}}$ (\emph{accomodating}) and $\mathcal{A}^{\text{NAC}}$ (\emph{non-accomodating})}
  \label{fig:gen_cog_hier}
\end{figure}
In order to support heterogeneous behavior models, the generalized dynamic cognitive hierarchy model consists of three layers of increasing sophistication of strategic reasoning, each of which can hold multiple behavior models. The three layers include: a) \emph{non-strategic}, where agents do not reason about the other agents' strategies, b) \emph{strategic}, where agents can reason about the strategies of other agents, and c) \emph{robust}, where agents not only reason about the strategies of other agents but also the behavior model \cite{wright2020formal} that other agents may be following (Fig. \ref{fig:gen_cog_hier}a). All of these layers operate in a setting of a dynamic game and we present the models and the solution concepts used in each layer in order. 
\subsection{Non-strategic layer}
Similar to level-0 models in the standard level-k reasoning \cite{camerer2004cognitive}, non-strategic agents form the base of the cognitive hierarchy in our model. However, in our case, we extend the behavior for the dynamic game. The main challenge of constructing the right level-0 model for a dynamic game is that it has to adhere to the formal constraints of non-strategic behavior, i.e. not reason over other agents' utilities \cite{wright2020formal}, while at the same time cannot be too elementary for the purpose of modeling human driving, which implies allowing for agents to change their behavior over time in the game based on the game state.\par
We propose that automata strategies, which were introduced as a way to address bounded rational behavior that results from agents having limited memory capacity to reason over the entire strategy space in a dynamic game \cite{rubinstein1986finite,marks1990repeated}, address the above problem by striking a balance between adequate sophistication and non-strategic behavior. To this end, we extend the standard level-k model for a dynamic game setting with level-0 behavior mediated by automata strategies (referred to as dLk($\mathcal{A}$) henceforth in the paper). In this framework, a level-0 agent has two reactive modes of operation, each reflecting a specific driving style modeled as automata strategies; \emph{accomodating} ($\mathcal{A}^{\text{AC}}$) and \emph{non-accomodating} ($\mathcal{A}^{\text{NAC}}$) (Fig. \ref{fig:gen_cog_hier}b). An agent $i$ playing $\mathcal{A}^{\text{AC}}$ as type $\gamma_{i}^{\text{AC}} \in \Gamma$ in state $\text{W}_{\text{AC}}$ randomize uniformly between trajectories belonging to \emph{wait} maneuver that have step safety utility at least $\gamma_{i}^{\text{AC}}$. If no such \emph{wait} trajectories are available to them, they move to state $\text{P}_{\text{AC}}$ and randomizes uniformly between trajectories belonging to \emph{proceed} maneuver. $\mathcal{A}^{\text{NAC}}$ is similar, but the states are reversed thereby resulting in a predisposition that prefers the proceed state. The switching between the states of the automata is mediated by the preference conditions $\phi_{i}^{\text{AC}}$ and $\phi_{i}^{\text{NAC}}$ shown in the equations below.
\small
\begin{linenomath*}
\begin{align}
    \phi_{i}^{\text{AC}}  &:=   \max\limits_{ \mathrm{T_i}(X_{t})|_{\text{W}}} u_{s,i}^{\Delta t_{p}}(\pi(X_{t})) \leqslant \gamma_{i}^{\text{AC}} 
    \label{eqn:phis_1}\\
    \phi_{i}^{\text{NAC}} &:= \max\limits_{\mathrm{T_i}(X_{t})|_{\text{P}}} u_{s,i}^{\Delta t_{p}}(\pi(X_{t})) > \gamma_{i}^{\text{NAC}}
    \label{eqn:phis_2}
\end{align}
\end{linenomath*}
\normalsize
$\phi_{i}^{\text{AC}}$ is true when there is at least one wait trajectory (and conversely, at least one proceed trajectory in the case of $\phi_{i}^{\text{NAC}}$) available whose step safety utility ($u^{\Delta t_{p}}_{s,i}$) at the current node $X_t$ is above $\gamma_{i}^{\text{AC}}$ (and $\gamma_{i}^{\text{NAC}}$ for $\phi_{i}^{\text{NAC}}$). Recall that agent types are in the range [-1,1] and are reflective of safety aspiration.\par
Along with switching between states, agents are also free to switch between the two automata; however, we leave open the question of modeling the underlying process and the causal factors. We envision them to be non-deterministic and a function of the agent's affective states, such as impatience, attitude, etc., which are an indispensable component of modeling driving behavior \cite{sagberg2015review}. For example, one can imagine a left turning vehicle approaching the intersection starting with an accommodating strategy, but along the game play, changes its strategy to non-accommodating on account of impatience or other endogenous factors. Although leaving open the choice of mode switching leads to the level-0 model in this paradigm to be partly descriptive rather than predictive, as we will see in the next section, such a choice does not compromise the ability of higher level agents' to form consistent beliefs about the level-0 agent and respond accordingly to their strategies. This richer model of level-0 behavior not only imparts more realism in the context of human driving, but also allows for level-0 agent to adapt their behavior based on the game situation in a dynamic game.
\subsection{Strategic layer}
The difference between strategic and non-strategic models is that the latter adhere to the two properties of \emph{other responsiveness}, i.e. reasoning over the utilities of the other agents, and \emph{dominance responsiveness}, i.e. optimizing over their own utilities \cite{wright2020formal}. Agents in the strategic layer in our framework adhere to the above two properties and we include three models of behavior, namely, level-k ($k \geqslant 1$) behavior based on the dLk($\mathcal{A}$) model and two types of bounded rational equilibrium behavior, i.e. SSPE (Safety satisfied perfect equilibria) and MSPE (Maneuver satisfied perfect equilibria). We note that the choice of models in the strategic layer is not exhaustive, however, we select popular game-theoretic models (level-k and equilibrium based) that have been used in the context of autonomous driving and address some of the gaps within the use of those models as secondary contributions. \par
For the strategic models to reason over the utilities of other agents and respond accordingly, the multi-objective utilities presented in Sec. \ref{sec:game_tree} need to be aggregated into a scalar utility. To that end, we use a lexicographic thresholding approach to aggregate the two multi-objective utilities \cite{LiChangjian19}, and furthermore, we connect the lexicographic threshold to an agent’s type $\gamma_{i}$. Specifically, the combined utility $u_{i}(\pi_{i}(h),\pi_{-i}(h))$ is equal to $u_{s,i}(\pi_{i}(h),\pi_{-i}(h))$, i.e., the safety utility when $u_{s,i}(\pi_{i}(h),\pi_{-i}(h)) \leqslant \gamma_{i}$, and otherwise $u_{i}(\pi_{i}(h),\pi_{-i}(h)) = u_{p,i}(\pi_{i}(h),\pi_{-i}(h))$, i.e., the progress utility. Based on the aggregation method, an agent would consider the progress utility of an action only when it's safety utility is greater than $\gamma_i$. This way of aggregation provides a natural connection between an agent's type and their risk tolerance. In the following sections, when we use the agent types in different behavior models, we also index the agent types with the specific models for clarity.\par
\subsection{dLk($\mathcal{A}$) model ($k \geqslant 1$ behavior)} A level-1 agent \footnote{We focus on $k=1$ behavior in this section as well as later in the experiments, but the best response behavior can be extended to $k>1$ similar to a standard level-k model.} believes that the population consists solely of level-0 agents, and generates a best response to those (level-0) strategies. In a dynamic setting, however, a level-1 agent has to update it's belief about level-0 agent based on observation of the game play. This means that in order to best respond to level-0 strategy, a level-1 agent should form a \emph{consistent} belief based on the observed  history $h_{t}$ of the game play about the type ($\gamma^{\text{AC}}_{-i},\gamma^{\text{NAC}}_{-i}$) a level-0 agent plays in each automaton.
\begin{defi}
\label{def:1}
A belief $\mathcal{B}_{l1} = \{\hat{\Upsilon}^{\text{AC}}_{l0}, \hat{\Upsilon}^{\text{NAC}}_{l0} \}$ of a level-1 agent at history $h_{t}$ is \emph{consistent} iff $\forall \gamma_{-i}^{\text{AC}} \in \hat{\Upsilon}^{\text{AC}}_{l0} $, $\pi_{-i}(X_{0})...\pi_{-i}(X_{t}) \in \mathcal{T}(\mathcal{A}^{\text{AC}};\gamma_{-i}^{\text{AC}},X_{0}..X_{t})$ and $\forall \gamma_{-i}^{\text{NAC}} \in \hat{\Upsilon}^{\text{NAC}}_{l0} $, $\pi_{-i}(X_{0})...\pi_{-i}(X_{t}) \in \mathcal{T}(\mathcal{A}^{\text{NAC}};\gamma_{-i}^{\text{NAC}},X_{0}..X_{t})$
where $\hat{\Upsilon}^{\text{AC}}_{l0} \subseteq \Gamma$, $\hat{\Upsilon}^{\text{NAC}}_{l0} \subseteq \Gamma$, and (dropping the superscripts) $\mathcal{T}(\mathcal{A};\gamma,X_{0}..X_{t})$ is the trace of the automaton, defined as the set of all valid sequence of actions generated by an level-0 agent playing an automaton $\mathcal{A}$ as type $\gamma$ and encountering the nodes $X_{0}..X_{t}$.
\end{defi}
Before the level-1 agent has observed any action by the level-0 agent, their estimates for $\hat{\Upsilon}^{\text{AC}}_{l0}$ and $\hat{\Upsilon}^{\text{NAC}}_{l0}$ is the default set of types, i.e. $\hat{\Upsilon}^{\text{AC}}_{l0} = \hat{\Upsilon}^{\text{NAC}}_{l0} = \Gamma$ with range 2 (recall $\Gamma = [-1,1]$). However, over time with more observations of level-0 actions, level-1 agent forms a tighter estimate, i.e. $|\max(\hat{\Upsilon}^{\text{AC}}_{l0}) - \min(\hat{\Upsilon}^{\text{AC}}_{l0})| \leqslant 2$, of level-0's type. The following
theorem formulates the set of consistent beliefs $\mathcal{B}_{l1}$ based on observed level-0 actions with history $h$ more formally.

\begin{theo}
\textit{For any consistent belief $\gamma_{-i}^{\text{AC}}$ and $\gamma_{-i}^{\text{NAC}}$, where $\gamma_{-i}^{\text{AC}} \in \hat{\Upsilon}^{\text{AC}}_{l0} $ and $\gamma_{-i}^{\text{NAC}} \in \hat{\Upsilon}^{\text{AC}}_{l0} $ the following inequalities hold true for any history $h$ of the game},
\begin{linenomath*}
 \begin{align*}
    \gamma_{-i}^{\text{AC}} & \begin{array}{l} < \min_{ h[\text{P}]} \max_{ \mathrm{T}_{-i}(X)|_{W}} u_{s,l0}^{\Delta_{t_p}}(\pi(X))  \\
     \geqslant \max_{ h[\text{W}]} \max_{ \mathrm{T}_{-i}(X)|_{W}} u_{s,l0}^{\Delta_{t_p}}(\pi(X)) \end{array}\\
     \gamma_{-i}^{\text{NAC}} & \begin{array}{l} \geqslant \max_{ h[\text{W}]} \max_{ \mathrm{T}_{-i}(X)|_{P}} u_{s,l0}^{\Delta_{t_p}}(\pi(X))  \\
      < \min_{ h[\text{P}]} \max_{ \mathrm{T}_{-i}(X)|_{P}} u_{s,l0}^{\Delta_{t_p}}(\pi(X))\end{array}
 \end{align*}
 \end{linenomath*}
\end{theo}

\noindent where $h[\text{P}]$ and $h[\text{W}]$ are set of game nodes where level-0 agent chose a proceed and wait maneuver respectively, and $T(X)|_{P}$,$T(X)|_{W}$ are the available trajectories at node $X$ belonging to the two maneuvers, proceed and wait, respectively.
\begin{proof}
There are two parts to the equation, one for $\gamma_{-i}^{\text{AC}}$ and another $\gamma_{-i}^{\text{NAC}}$. We prove the bounds of $\gamma_{-i}^{\text{AC}}$ corresponding to automata $\mathcal{A}^{\text{AC}}$, and the proof for $\gamma_{-i}^{\text{NAC}}$ corresponding to automata $\mathcal{A}^{\text{NAC}}$ follows in an identical manner.\par
By construction of automata $\mathcal{A}^{\text{AC}}$, proceed trajectories are only generated in state $\text{P}_{\text{AC}}$, which  follows one of the three transitions $\{ \mathrm{inp.} | \text{P}_\text{AC} | \text{W}_\text{AC} \}\xrightarrow{\neg \phi^{\text{AC}}_{i}(X_{t})} \text{P}_\text{AC}$. Therefore, $\forall X_{t} \in h[\text{P}]$, $\phi_{i}^{\text{AC}} \coloneqq \bot$ for the transition to happen. Based on eqn. 1, this means that $\forall X_{t} \in h[\text{P}]$, 
\begin{align}
\gamma_{i}^{\text{AC}} &< \max\limits_{ \mathrm{T_{-i}}(X_{t})|_{\text{W}}} u_{s,i}^{\Delta t_{p}}(\pi(X_{t}))    \tag{A.1}
\end{align}

Since $\gamma_{i}^{\text{AC}}$ stays constant throughout the play of the game, for $\gamma_{i}^{\text{AC}}$ to be consistent for the set of all nodes $X_{t} \in h[\text{P}]$, the lower bound (i.e., at least as high to be true for all nodes $X_{t} \in h[\text{P}]$) of $\gamma_{i}^{\text{AC}}$ based on eqn A.1 is
\begin{align}
   \gamma_{i}^{\text{AC}} &< \min_{ h[\text{P}]} \max_{ \mathrm{T_{-i}}(X_{t})|_{W}} u_{s,l0}(\pi(X)) \tag{A.2} 
\end{align}
\par
Similarly, by construction of automata, wait trajectories are only generated in state $\text{W}_\text{AC}$, which  follows one of the three transitions $\{ \mathrm{inp.} | \text{P}_\text{AC} | \text{W}_\text{AC} \}\xrightarrow{ \phi^{\text{AC}}_{i}(X_{t})} \text{W}_\text{AC}$. Therefore, $\forall X_{t} \in h[\text{W}]$, $\phi_{i}^{\text{AC}} \coloneqq \top$ for the transition to happen. Based on eqn. 1, this means that $\forall X_{t} \in h[\text{W}]$, 
\begin{equation}
\gamma_{i}^{\text{AC}} \geqslant \max\limits_{\mathrm{T_{-i}}(X_{t})|_{\text{P}}} u_{s,i}^{\Delta t_{p}}(\pi(X_{t}))    \tag{A.3} 
\end{equation}
and the upper bound (i.e., at least as low to be true for all nodes $X_{t} \in h[\text{W}]$) of $\gamma_{i}^{\text{AC}}$ based on eqn A.3 $\forall X_{t} \in h[\text{W}]$
\begin{equation}
   \gamma_{i}^{\text{AC}} \geqslant \max_{ h[\text{W}]} \max_{ \mathrm{T}(X_{t})|_{W}} u_{s,l0}(\pi(X)) \tag{A.4} 
\end{equation}
Since $h[\text{P}] \bigcap h[\text{W}] = \varnothing$ and $h = h[\text{P}] \bigcup h[\text{W}]$, equations A.2 and A.4 in conjunction proves the case for $\gamma_{i}^{\text{AC}}$ bounds.
\par
The proof for $\gamma_{i}^{\text{NAC}}$ follows in the identical manner as $\gamma_{i}^{\text{AC}}$, but with the condition reversed based on the P and W states.
\end{proof}
The above theorem formalizes the idea that looking at maneuver choices made by a level-0 agent at each node in the history, as well as the range of step safety utility at that node for both maneuvers (recall that preference conditions of the automata are based on step safety utility), a level-1 agent can calculate ranges for $\gamma_{-i}^{\text{AC/NAC}}$ from Eqns \ref{eqn:phis_1} and \ref{eqn:phis_2}, for which the observed actions were consistent with each automata. With respect to the set of consistent belief $\mathcal{B}_{l1}$ about level-0 agent's strategy, level-1 agent now needs to generate a best response that is consistent with $\mathcal{B}_{l1}$. Dropping the AC/NAC superscripts Let $\pi(X_{t};\mathcal{A},\hat{\Upsilon}_{l0}) = \{\pi(X_{t};\mathcal{A},\gamma_{-i}); \forall \gamma_{-i} \in \hat{\Upsilon}_{l0} \}$ be the union of all actions when the automata $\mathcal{A}$ is played by the types in $\hat{\Upsilon}_{l0}$, then $\Pi_{l0}(X_{t},\mathcal{B}_{l1}) =  \pi(X_{t};\mathcal{A}^{\text{AC}},\hat{\Upsilon}^{\text{AC}}_{l0}) \bigcup \pi(X_{t};\mathcal{A}^{\text{NAC}},\hat{\Upsilon}^{\text{NAC}}_{l0})$ is the set of all actions that level-0 agent can play based on level-1's consistent belief $\mathcal{B}_{l1}$. The response to those actions by level-1 agent (indexed as $i$) is as follows.
\small
\begin{linenomath*}
\begin{align}
    \pi_{i}(h_{t};\mathcal{B}_{l1}) = &\argmax_{\substack{\mathrm{T}_{i}(X_{t}) \\  \Pi_{l0}(X_{t},\mathcal{B}_{l1})}} u_{i}(\pi_{i}(h),\pi_{-i}(X_{t})|\gamma_{i}^{l1})
\label{eqn:l1_resp}
\end{align}
\end{linenomath*}
\normalsize
where $\gamma_{i}^{l1} \in \Gamma$ is the type of the level-1 agent, $ \pi(h) \in \mathrm{T}_{i}(X_{t})$ is a valid trajectory that can be generated at the node with history $h$, and $\pi_{-i}(X_{t}) \in \Pi_{l0}(X_{t},\mathcal{B}_{l1})$ is the other agent's trajectory. Note that the strategy of the level-1 agent, unlike level-0 agent depends on the history $h_{t}$ instead of just the state of the node $X_{t}$; since the history influences the belief $\mathcal{B}_{l1}$ which in turn influences the response.\par
The above model of best response can be extended similarly for $k > 1$ just like in standard level-k models. However, we restrict our focus for $k$ up to 1 mainly due to following reasons: i) aspects of higher level thinking are already incorporated in the equilibrium based models; ii) in other domains, we often see diminishing returns with higher k values with respect to empirical data \citep{wright2010beyond}.
\subsection{Equilibrium models} Along with the level-k (k$\geqslant 1$) behavior, another notion of strategic behavior that have been proposed as a model of behavior planning in AV \cite{pruekprasert2019decision, schwarting2019social} are based on an equilibrium. However, when it comes to modelling human driving behavior, an equilibrium model needs to accommodate bounded rational agents in a principled manner, and ideally should provide a reasonable explanation of the origin of the bounded rationality. Based on the idea that drivers are indifferent as long as an action achieves their own subjective safety aspiration threshold \cite{lewis2012testing}, we use the idea of \emph{satisficing} as the main framework for bounded rationality in our equilibirum models \cite{stirling2003satisficing}. Specifically, we develop two notions of satisficing; one based on safety satisficing (SSPE), where agents choose actions close to the Nash equilibria as long as the actions are above their own safety aspiration threshold, and another based on maneuver satisficing (MSPE), where agents chose actions close to the Nash equilibria as long as the actions are of the same high-level maneuver as the optimal action.
\subsubsection{Safety-satisfied perfect equilibrium (SSPE).} The main idea behind satisficing is that a bounded rational agent, instead of always selecting the best response action, selects a response that is \emph{good enough}, where \emph{good enough} is defined as an aspiration level where the agent is indifferent between the optimal response and the response in question. In the case of Safety-satisfied perfect equilibrium (SSPE), we define a response good enough for agent $i$ if the response is above their own safety aspiration threshold as determined by their type. The goal of this model is to let agents chose their action based on the subgame perfect Nash equilibrium of the game, however, also allow for selection of actions close to the equilibrium that are safe enough based on their own safety aspiration level. Therefore, in response to other agents' equilibrium action, any action that is safer than either their own equilibrium action or their own safety aspiration level $\gamma_i$ is part of the solution. A more formal definition is as follows
\begin{defi}
\label{def:2}
A strategy $\sigma;(\gamma_{i}^{\text{SS}},\gamma_{-i}^{\text{SS}}) : \prod \limits_{\forall i \in N} \pi_{i}(h)$, is in safety satisfied perfect equilibria for a combination of agent types $(\gamma_{i}^{\text{SS}},\gamma_{-i}^{\text{SS}}) \in \Gamma^{N}$ if for every history $h$ of the game and $\forall i \in N$
\begin{linenomath*}
\begin{align*}
    u_{s,i}(\pi_{i}(h),\pi_{-i}^{*}(h))  \geqslant \min\{u_{s,i}^{*}(h), \gamma_{i}^{\text{SS}}\}
\end{align*}
\end{linenomath*}
\noindent where $\sigma^{*};(\gamma_{i}^{\text{SS}},\gamma_{-i}^{\text{SS}}) : \prod \limits_{\forall i \in N} \pi_{i}^{*}(h)$, is a subgame perfect Nash equilibrium in pure strategies of the game for agents with type $(\gamma_{i}^{\text{SS}},\gamma_{-i}^{\text{SS}})$ and $u_{s,i}^{*}(h) = u_{s,i}(\pi^{*}_{i}(h),\pi^{*}_{-i}(h))$. \footnote{We calculate the SPNE using backward induction with the combined utilities (lexicographic thresholding) in a complete information setting  where agent types are known to each other.}
\end{defi}
Based on the above definition, if the safety utility of the best response of agent $i$ to agent $-i$'s subgame perfect Nash equilibrium (SPNE) strategies at history $h$ is less than agent $i$'s own safety threshold as expressed by their type $\gamma_{i}^{\text{SS}}$, then the SSPE response is any trajectory that matches the safety utility of the SPNE response. However, if the SPNE response is higher than their safety threshold, then any suboptimal response that has safety utility higher than $\gamma_{i}^{\text{SS}}$ is a \emph{satisfied} response, and thus in SSPE.
\subsubsection{Maneuver-satisfied perfect equilibrium (MSPE).} This model of satisficing is based on the idea that agents may be indifferent between actions that belong to the maneuver corresponding to the equilibrium trajectory. For example, at any node, if the equilibrium trajectory belongs to \emph{wait} maneuver, then all trajectories belonging to \emph{wait} maneuver will be in MSPE. However, there are additional constraints that need to be imposed on this manner of action selection. In order to avoid selection of trajectories that belong to equilibrium maneuver but have lower utility than a non-equilibrium maneuver, we add the constraint that the utility of a selected trajectory has to be strictly higher than that of any non-equilibrium maneuver's trajectory. A typical example for the need of this constraint include situations where the equilibrium trajectory corresponds to \emph{wait} maneuver, but selecting a trajectory that is akin to a rolling stop, which although falls under wait maneuver, is worse than executing a proceed trajectory, i.e., a non-equilibrium maneuver. Therefore, such trajectories should be excluded from the MSPE solution set. A formal definition follows.
\begin{defi}
\label{def:3}
A strategy $\sigma;(\gamma_{i}^{\text{MS}},\gamma_{-i}^{\text{MS}}) : \prod \limits_{\forall i \in N} \pi_{i}(h)$, is in maneuver satisfied perfect equilibria for a combination of agent types $(\gamma_{i}^{\text{MS}},\gamma_{-i}^{\text{MS}}) \in \Gamma^{N}$ if for every history $h$ of the game and $\forall i \in N$, $m(\pi_{i}(h)) = m(\pi_{i}^{*}(h))$ and 
\small
\begin{linenomath*}
\begin{align*}
    u_{s,i}(\pi_{i}(h),\pi_{-i}^{*}(h)) &> \max_{\pi^{'}_{i}(h) \in \mathrm{T}_{i}(X) \setminus m^{*} } u_{i}(\pi^{'}_{i}(h),\pi_{-i}^{*}(h))
\end{align*}
\end{linenomath*}
\normalsize
\noindent where $\sigma^{*};(\gamma_{i}^{\text{MS}},\gamma_{-i}^{\text{MS}}) : \prod \limits_{\forall i \in N} \pi_{i}^{*}(h)$, is a subgame perfect equilibrium in pure strategies of the game  with agent types $(\gamma_{i}^{\text{MS}},\gamma_{-i}^{\text{MS}})$, and $\mathrm{T}_{i}(X) \setminus m^{*} = \{\pi_{i}(h) : m(\pi_{i}(h)) \neq m(\pi_{i}^{*}(X))\}$ or in other words, the set of available trajectories at node $X$ that do not belong to the maneuver corresponding to the equilibrium trajectory $m(\pi_{i}^{*}(h))$ and $X$ is the last node in the history $h$.
\end{defi}
\subsection{Robust layer}
While the presence of multiple models in the strategic layers allow for a population of heterogeneous reasoners, an agent following one of those models still has specific assumptions about the reasoning process of other agents, e.g. level-1 agents believing that the population consists of level-0 agents and equilibrium responders believing that other agents adhere to a common knowledge of rationality. Based on the position that a planner for AV should not hold such strict assumptions, we develop the robust layer. What differentiates the robust layer from the strategic layer is that along with the two properties of other responsiveness and dominance responsiveness for the strategic layer, agents in the robust layer also adhere to the property of \emph{model responsiveness}, i.e., the ability to reason over the behavior models of other agents. This gives them the ability to reason about (forming beliefs about and responding to) a population of different types of reasoners including strategic, non-strategic, as well as agents following different models within each layer. The overall behavior of a robust agent can be broken down into three sequential steps as follows.\\
\noindent{\textbf{i. Type expansion:}} Since the robust agent not only has to reason over the types of other agents, but also the possible behavior models, we augment the initial set of agent types $\Gamma$ that were based on agents' safety aspiration with the corresponding agent models. Let $\Gamma^{+} : \mathcal{M} \times \Gamma$ be the augmented type of an agent, where  $\mathcal{M}$ is the set of models presented earlier, i.e., \{accomodating, non-accomodating, level-1, SSPE, MSPE\} and $\Gamma$ is the (non-augmented) agent type  agent type ($\gamma_{i}^{\text{AC}}, \gamma_{i}^{\text{NAC}}, \gamma_{i}^{\text{l1}}, \gamma_{i}^{\text{SS}}, \gamma_{i}^{\text{MS}}$) within each model.\par
\noindent{\textbf{ii. Consistent beliefs:}} Similar to strategic agents, based on the observed history $h$ of the game, a robust agent forms a belief $\upbeta_h \subseteq \Gamma^{+}$ such that the observed actions of the other agent in the history $h$ is consistent with the augmented types (i.e., model as well as the agent type) in $\upbeta_h$. The process of checking whether a history is consistent with a combination of a model and agent type was already developed earlier for two non-strategic models (Def. \ref{def:1}). For level-1 models, the history is consistent if at each node in history, the response of the other agent adheres to equation \ref{eqn:l1_resp}; and for the equilibrium models, a history $h$ is consistent if based on definitions \ref{def:2} and \ref{def:3} for SSPE and MSPE respectively, the actions are along the equilibrium paths of the game tree. Assuming that in driving situations agents behave truly according to their types, $\upbeta_h$ is then constructed as an union of all the consistent beliefs for each model.\par
\noindent{\textbf{iii. Robust response:}} The idea of a robust response to heterogeneous models is along the lines of the robust game theory approach of \cite{aghassi2006robust}. The belief set $\upbeta_h$ represents the uncertainty over the possible models the other agents' may be using along with the corresponding agent types within those models. A robust response to that is the optimization over the worst possible outcome that can happen with respect to that set. Eqn. \ref{eqn:robust_resp} formulates this response of a robust agent playing as agent $i$.
\small
\begin{linenomath*}
\begin{equation}
\pi_{i}(h;\upbeta_h) = \argmax_{\text{T}_{i}(X)} \inf_{\forall \beta \in \upbeta_{h}} \max_{\forall \pi_{-i}(h;\beta)} u_{i}(\pi(h), \pi_{-i}(h;\beta);\gamma_{i}^{\text{R}})
\label{eqn:robust_resp}
\end{equation}
\end{linenomath*}
\normalsize
where $\pi_{-i}(h;\beta)$ are the possible actions of the other agent based on the augmented type $\beta \in \upbeta_{h}$ and $\gamma_{i}^{\text{R}} \in \Gamma$ is the robust agent's own type. In this response, the minimization happens over the agent types (inner $\inf$ operator), rather than over all the actions $\pi_{-i}(h)$ as is common in a maxmin response. Since driving situations are not, in most cases, purely adversarial, this is a less conservative, yet robust, response compared to a maxmin response.
\par
\section{Experiments and evaluation}
\begin{figure}[!ht]
         \centering
         \begin{subfigure}[b]{\linewidth}
         \centering
          \includegraphics[width=\linewidth]{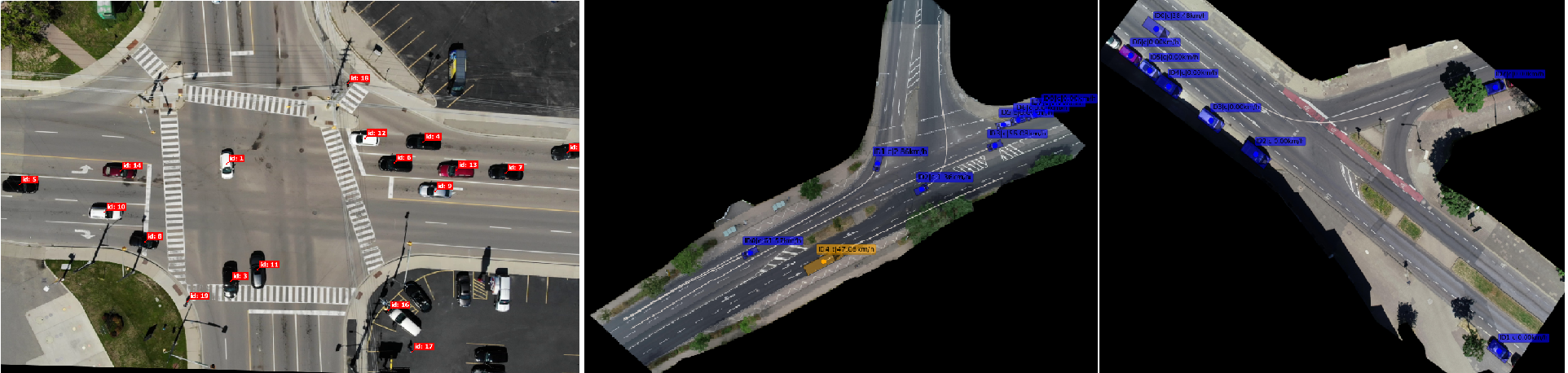}
          \caption{Snapshot of naturalistic datasets (WMA and inD)}
          \label{fig:nat_snap}
        \end{subfigure}
        \begin{subfigure}[b]{\linewidth}
         \centering
          \includegraphics[width=.7\linewidth]{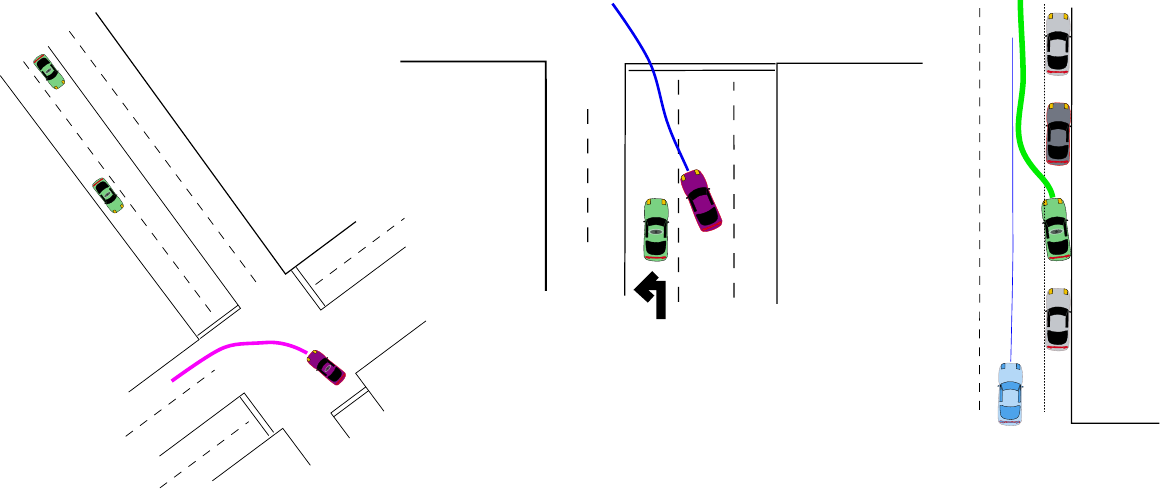}
          \caption{Simulation of critical scenarios: intersection clearance, merge before intersection, parking pullout.}
          \label{fig:crit_snap}
        \end{subfigure}
    \label{fig:eval_set}
    \caption{Evaluation setup based on naturalistic datasets and simulation scenarios.}
\end{figure}
In this section we present the evaluation of the models under two different experiment setups. First, we compare the models with respect to large naturalistic observational driving data using a) the Intersection dataset from the Waterloo multi-agent traffic dataset (WMA) recorded at a busy Canadian intersection \cite{sarkar2021solution} available at \url{http://wiselab.uwaterloo.ca/waterloo-multi-agent-traffic-dataset/}, and b) the inD dataset recorded at intersections in Germany \cite{inDdataset} (Fig. \ref{fig:nat_snap}). From both datasets, which include around 10k vehicles in total, we extract the long duration unprotected left turn (LT) and right turn (RT) scenarios, and instantiate games between left (and right) turning vehicles and oncoming vehicles with $\Delta t_{h}=6$s and $\Delta t_{p}=2$s, resulting in a total of 1678 games. The second part of the evaluation is based on simulation of three critical traffic scenarios derived from the NHTSA pre-crash database \cite{najm2007pre}, where we instantiate agents with a range of safety aspirations as well as initial game states, and evaluate the outcome of the game based on each model. All the games in the experiments are 2 agent games with the exception of one of the simulation of critical scenario (intersection clearance), which is a 3 agent game. Code and supplementary videos are available at \url{https://git.uwaterloo.ca/a9sarkar/repeated_driving_games}. \par
\noindent \textbf{Baselines.} We select multiple baselines depending on whether a model is strategic or non-strategic. For non-strategic models, we compare the automata based model with a maxmax model, shown to be most promising from a set of alternate elementary models with respect to naturalistic data \cite{sarkar2021solution}. For the strategic models (level-1 in dLk($\mathcal{A}$), SSPE, MSPE), we select a QLk model used in multiple works within the context of autonomous driving \cite{li2018game, tian2018adaptive, tian2021anytime, li2019decision}. We use the same parameters used in \cite{tian2021anytime} for the precision parameters in the QLk model.
\subsubsection{Naturalistic data}
We evaluate the model performance on naturalistic driving data based on match rate, i.e. the number of games where the observed strategy of the human driver matched a model's solution divided by the total number of games in the dataset.  More formally, let $D$ be the set of games in the dataset, an indicator function $I:D \rightarrow \{0,1\}$ is 1 if in the game $g$, there exists a combination of agent types ($\gamma_{i},\gamma_{-i}$), such that the observed strategy is in the set of strategies as predicted by the model, or 0 otherwise. The overall match rate of a model is given by $\sum_{\forall g \in D} I/|D|$. QLk (baseline) models being mixed strategy models, we count a match if the observed strategy is assigned a probability of $\geqslant 0.5$.
\begin{table}[!t]
\tiny
\centering
\begin{tabular}{P{1.25cm}P{1cm}P{1cm}P{1cm}P{1cm}}
    \toprule
    &\multicolumn{2}{c}{WMA}&\multicolumn{2}{c}{inD}\\
    &LT (1103)&RT (311)&LT (187)&RT (77)\\
    \midrule
    maxmax&0.33438 (0.02)&0.43023 (0.27)& 0.37975 (-0.2)&0.43506 (0.1)\\
    AC&0.82053 (0.82)&0.90698 (0.84)&0.92089 (0.75)&0.81818 (0.79)\\
    NAC&0.17947 (0.87)&0.09302 (0.82)&0.07911 (0.88)&0.18182 (0.85)\\
    \midrule
    QLk($\lambda$=1)&0.18262 (-0.07)&0.43265 (0.33)&0.37658 (0.37)&0.43506 (-0.1)\\
    QLk($\lambda$=0.5)&0.34131 (-0.03) &0.43023 (0.26)&0.37658 (0.37)&0.43506 (-0.1)\\
    dLk($\mathcal{A}$)&0.5529 (0.19)&0.65449 (0.3)&0.51266 (0.51)&0.53247 (-0.4)\\
    SSPE&0.69144 (-0.84)&0.90033 (-0.94)&0.6962 (-0.86)&0.53247 (-0.85)\\
    MSPE&0.30479 (-0.1)&0.44518 (-0.13)&0.21519 (0.21)&0.27273 (-0.7)\\
    \midrule
    Robust&0.56045 (0.20)&0.66944 (0.34)&0.51582 (0.51)&0.53247 (-0.4)\\
    \bottomrule
\end{tabular}
\small
\caption{Overall match rate of the models for each dataset and scenario. Mean agent type ($\gamma$) noted in parenthesis. LT: Left turn, RT: Right turn. Number of games noted in the header. }
\label{tab:model_accu}
\end{table}
Table \ref{tab:model_accu} shows the match rate of each model for each dataset and scenario. It also shows in parenthesis the mean $\gamma_{i}$, i.e. the agent type value for each model when the strategy matched the observation. Since the agent types are a free parameter within each model, this gives the models flexibility to match the observed action to a driver's safety aspiration level. The overall numbers in the table show that there is variation both with respect to match rate as well as the matched safety aspiration level (agent type) for a given model. We observe that the match rate, especially for WMA dataset, is better for right turning scenarios than for left turning ones. The models sometimes find harder to match a consistent safety aspiration level at left turns in WMA when vehicles exhibit impatient behavior by creeping forward. Notice that the difference is less stark in the inD dataset because in inD (Fig. 3a), the LT vehicle is still in the turn lane at point of initiation (i.e., before the stopline), and therefore has less incentive to take a risk and creep. \par
A major takeaway is that for non-strategic models, automata models show much higher match rate thereby reflecting high alignment with human driving behavior compared to the maxmax model. In fact, as we can see from the table that the entries for AC and NAC sum up to 1. Combination of AC and NAC although being non-strategic, can capture all observed driving decisions in the dataset, which indicate that automata models are very well suited for modelling level-0 behavior in a dynamic game setting for human driving. The performance of an accommodating model is not very surprising since for left and right turning scenarios, most drivers generally give way to oncoming vehicles. For the strategic models, dLK($\mathcal{A}$) and SSPE models show better performance than QLk and MSPE models. However, when we compare the mean agent types, we observe that when SSPE strategies matches the observation, it is based on agents with very low safety aspiration (reflected in low mean agent type values). If we assume that the population of drivers on average have moderate safety aspiration, say in the range [-0.5,0.5] (estimating the true distribution is out of the current scope), dLk($\mathcal{A}$) is a more reasonable model of strategic behavior compared to SSPE. We include the robust model comparison for the sake of completeness (and it shows performance comparable to dLk($\mathcal{A}$) model), but as mentioned earlier, robust model is a model of behavior planning for an AV, and therefore ideally needs to be evaluated on criteria beyond just comparison to naturalistic human driving, which we discuss in the next section.
\subsubsection{Critical scenarios}
While evaluation based on a naturalistic driving datasets helps in the understanding of how well a model matches human driving behavior, in order to evaluate the suitability of a model for behavior planning of an AV, the models need to be evaluated on specific scenarios that encompass the operational design domain (ODD) of the AV \cite{ilievski2020wisebench}. Since the models developed in this paper are not specific to an ODD, we select three critical scenarios from ten most frequent crash scenarios in the NHTSA pre-crash database \cite{najm2007pre}. \par
\textbf{Intersection clearance (IC):} Left turn across path (LTAP) scenario where the traffic signal has just turned from green to yellow at the moment of the game initiation. There is a left turning vehicle is on the intersection and two oncoming vehicles from the opposite direction close to the intersection who may chose to speed and cross or wait for the next cycle. The expectation is that the left turning vehicle should be able to clear the intersection by the end of the game horizon without crashing into either oncoming vehicles, and no vehicles should be stuck in the middle of the intersection.\par
\textbf{Merge before intersection (MBI):} Merge scenario where a left-turning vehicle (designated as the merging vehicle) finds itself in the wrong lane just prior to entering the intersection, and wants to merge into the turn lane in front of another left-turning vehicle (designated as the on-lane vehicle). The expectation is that the on-lane vehicle should allow the other vehicle to merge.\par
\textbf{Parking pullout (PP):} Merge scenario where a parked vehicle is pulling out of a parking spot and merges into traffic while there is a vehicle coming along the same direction from behind. The expectation is that the parked vehicle should wait for the coming vehicle to pass before merging into traffic.\par
For each scenario, we run simulations with a range of approach speeds as well as all  combination of agent types from the set of agent types $\Gamma = \{-1, -0.5, 0, 0.5, 1\}$.
\begin{figure}[!t]
         \centering
         \includegraphics[width=\linewidth]{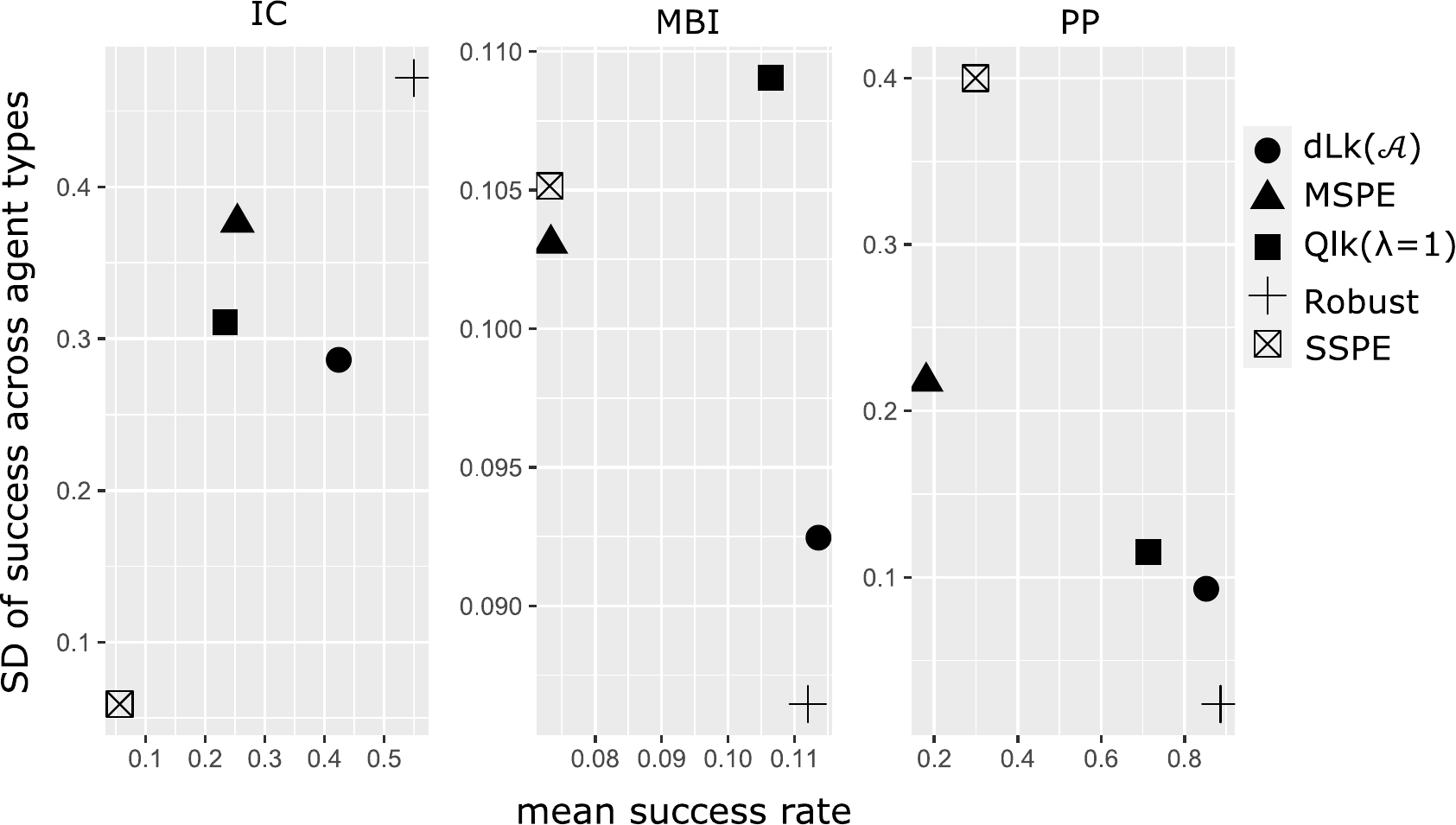}
         \caption{Mean and SD of success for each model in each scenario across all agent types.}
         \label{fig:type_succ}
\end{figure}
One way to compare the models is to evaluate them based on the mean success rate across all initiating states and agent types. Fig. \ref{fig:type_succ} shows the mean success rate (success defined as the desired outcome based on expectation defined in the description for each scenario) for all the strategic and robust models. We see that the mean success rate of the robust and dLk(A) model is higher compared to the equilibrium models or the QLk model. However, this is only part of the story. With varying initiation conditions, it may be harder or easier for a model to lead to a successful outcome. For example, in the parking pullout scenario a vehicle with low safety aspiration coming at a higher speed is almost likely to succeed in all models when facing a parked vehicle with high safety aspiration at zero speed. Therefore, to tease out the stability of models across different safety aspirations (i.e. agent type combinations), Fig. \ref{fig:type_succ} also plots on y-axis, the standard deviation of the mean success rate across different agent types. Ideally, a model should have a high success rate with low SD across types indicating that with different combinations of agent type population (from extremely low safety aspiration to very high), the success rate stays stable. As we see from Fig. \ref{fig:type_succ}, the robust and dLk(A) models are broadly in the ideal lower right quadrant (high mean success rate low SD) for parking pullout and merge before intersection scenarios. For IC scenario, however, we observe that the success rate comes at a price of high SD (for all models) as indicated by the linearly increasing relation between mean success rate and its SD across agent types. This means that the success outcomes are skewed towards a specific combination of agent types; specifically, the case where the left turning vehicle has low safety aspiration. It is intuitive to imagine that in a situation like IC, agents with high safety aspiration would be stuck in the intersection instead of being able to navigate out of the intersection quickly.\par
Finally, the failure of models to achieve the expected outcome can also be due to a crash (minimum distance gap between trajectories $\leqslant 0.1$m) instead of an alternate outcome (e.g. getting stuck in the intersection). In all the simulations, for the parking pullout and intersection clearance we did not observe a crash for any of the models. However, for the merge before intersection, on account of starting at a more riskier situation than the other two in terms of chance of a crash, the crash rate (ratio of crashes across all simulations) for the models across all initial states and agent types were as follows: (dlk($\mathcal{A}$): 0.052, MSPE: 0.022, SSPE: 0.007, QLk(level-1): 0.026, Robust: 0.053). QLk (level-1) demonstrates conservative behavior primarily due to maxmax behavior of level-0 agent, where always best responding to maxmax behavior ends up being more conservative than best-responding to diverse models by belief updating (as in robust and dlk(A)). This is reflected in the lower crash rate for QLk model compared to Robust. Equilibrium models lead to reduced collision rate compared to dlk(A) and robust, likely a result of working under a complete information setting where there is no scope for misinterpreting the other agents’ type. In terms of success rate, MBI also shows lower success rate for all models. This is mainly because the situation (low distance gap between vehicle during game initiation) is setup as such that for most models the optimal strategy profile is merging vehicle waiting for the on lane vehicle to pass. In fact, the WMA dataset had one instance of MBI scenario, which resulted in failure. Similarly, we note that collisions are also observed only in the MBI scenario.\par
Overall, whether or not an AV planner can succeed in their desired outcome depends on a variety of factors, such as, the assumption the vehicle and the human drivers hold over each other, the safety aspiration of each agent, as well as the specific state of the traffic situation. The analysis presented above helps in quantifying the relation between the desired outcome and the criteria under which it is possible.
\section{Conclusion}
We developed a unifying framework of modeling human driving behavior and strategic behavior planning of AVs that supports heterogeneous models of strategic and non-strategic reasoning. The model consists of three layers of increasing ability for strategic reasoning, and each layer can hold multiple behavior models. For the non-strategic layer, we also developed a model of level-0 behavior for level-k type models through the use of automata strategies (dLk($\mathcal{A}$)) that is suitable as a non-strategic model for the context of modeling driving behavior. The evaluation on two large naturalistic datasets shows that a combination of a rich level-0 behavior can capture most of the driving behavior as observed in the dataset. On the other hand, for the problem of behavior planning, with the awareness that there can be different types of reasoners in the population, an approach of robust response to heterogeneous behavior models is not only effective, but also is stable across a population of drivers with different levels of risk tolerance.

\begin{quote}
\begin{small}
\bibliography{main}
\end{small}
\end{quote}

\end{document}